\documentclass[orange]{isrpaper}

\IfFileExists{binhex.tex}{%
  
  \RequirePackage{newtxtext,newtxmath}
  \ifcsname XeTeXrevision\endcsname\usepackage{fontspec}\fi
}{%
  \RequirePackage{mathptmx}
}

\usepackage{multirow}
\usepackage{longtable}
\usepackage{tabularx}
\usepackage{subcaption}
\usepackage{algorithm}
\usepackage{algpseudocode}
\usepackage{csquotes}

\usepackage{orcidlink}

\usepackage[
    backend=biber,
    style=authoryear,
    natbib=true,
    doi=false,
    url=false,
    isbn=false,
    maxbibnames=99,
    maxcitenames=2,
    uniquelist=false,
]{biblatex}

\AtEveryBibitem{%
   \clearfield{Title}%
   \clearfield{eprint}%
   \clearfield{note}%
}

\renewcommand{\citet}[1]{\textcite{#1}}
\renewcommand{\citep}[1]{\parencite{#1}}

\title{Probabilistic Monocular Biomechanical Reconstruction via Guided Flow Matching}

\author{
R. James Cotton\,\orcidlink{0000-0001-5714-1400}\,$^{1,2}$, Fabian H. Sinz\,\orcidlink{0000-0002-1348-9736}\,$^{3}$}

\affiliations{
$^{1}$Shirley Ryan AbilityLab \quad $^{2}$Department of Physical Medicine and Rehabilitation, Northwestern University \quad $^{3}$Department of Computer Science and Campus Institute Data Science, University of G\"ottingen, G\"ottingen, Germany}

\paperdate{October 2026}

\correspondence{rcotton@sralab.org}

\paperabstract{Reconstructing 3D joint kinematics from a single video is ill-posed due to depth--stature ambiguity, limb occlusion, and missing uncertainty estimates. We introduce the Biomechanical Flow Model (BFM), a continuous-time flow-matching prior over subject anthropometry, anatomical joint-angle trajectories, and virtual markers, trained on multi-camera clinical recordings from 565 participants and steered toward monocular 2D keypoints at test time without paired video training. In a head-to-head comparison on the same clinical trials against a state-of-the-art deterministic trajectory optimizer fitting the identical skeleton, BFM achieves comparable joint-angle accuracy across walking and timed-up-and-go (TUG). Across held-out control, prosthetic, and neurologic participants, BFM reaches 3.91\ensuremath{^\circ} and 4.97\ensuremath{^\circ} mean absolute joint-angle error on walking and TUG while providing credible intervals that bridge temporal and whole-limb occlusions. Where keypoints are hidden, the intervals widen with the error and keep their coverage near nominal. Supplying individual anthropometric calibration removes most monocular pelvis depth error, whereas testing on out-of-distribution pediatric crouch gait produces biased, overconfident extension predictions, highlighting the need for pathologically diverse movement priors.}

\begin{document}

\maketitle

\section{Introduction}

Clinical movement analysis relies on 3D joint kinematics to quantify impairments, guide treatment, and track recovery. Marker-based capture and multi-camera markerless systems recover accurate 3D joint trajectories \parencite{kanko_concurrent_2021, uhlrich_opencap_2023, cotton_differentiable_2025}, yet require calibrated multi-camera laboratories impractical for routine clinics and homes. Single-camera video removes that barrier for sagittal and spatiotemporal screening \parencite{kidzinski_deep_2020, stenum_clinical_2024}, but recovering full 3D kinematics from one view remains ill-posed: participant stature and camera distance trade off along viewing rays, limbs occlude one another during turns and transfers, and transverse rotations produce subtle image cues.

Deterministic monocular trajectory optimizers \parencite{peiffer_portable_2025, gilon_opencap_2026, cotton_cambr_2026} and skeletal regressors \parencite{xia_reconstructing_2025, koleini_biopose_2025, cotton_biomechanical_2026} return a single point estimate without calibrated confidence intervals, failing silently when observations become ambiguous. Multi-camera variational inference provides calibrated intervals when views overlap \parencite{cotton_biomechanical_2025, donahue_calibrated_2026}. For monocular video, Platypose \parencite{pierzchlewicz_platypose_2024} built upon energy-guided diffusion (EGG) \parencite{pierzchlewicz_energy_2023} to steer a pretrained motion prior with a 2D reprojection energy, producing calibrated multi-hypothesis 3D joint positions without paired 2D--3D retraining. However, Platypose operated on unconstrained 3D joint coordinates rather than anatomical joint angles and participant anthropometry.

We introduce the Biomechanical Flow Model (BFM), a statistical flow-matching prior \parencite{lipman_flow_2023, tong_improving_2024} over anatomical joint-angle trajectories, virtual markers, and participant anthropometry (Figure~\ref{fig-overview}). At test time, differentiable forward-kinematic reprojection gradients steer the learned velocity field toward monocular 2D keypoints without retraining, generating an ensemble of posterior trajectory hypotheses per trial. Our contributions are:

\begin{enumerate}
\item \textbf{Biomechanical and anthropometric flow prior:} A continuous-time flow-matching prior over skeletal joint-angle trajectories and body-scaling parameters, trained with minibatch optimal-transport coupling \parencite{tong_improving_2024} on a diverse clinical and control cohort so anthropometry can either be inferred from video or supplied from individual calibration.
\item \textbf{Zero-shot forward-kinematic guidance:} A guided flow sampler steering trajectories from monocular 2D keypoints via differentiable forward kinematics, camera alignment, and skeletal projection.
\item \textbf{Clinical accuracy, uncertainty calibration, and generalization limits:} Validation on unseen participants demonstrating joint-angle accuracy comparable to state-of-the-art deterministic trajectory optimization, near-calibrated credible intervals whose coverage stays near nominal when keypoints are hidden, plausible infilling across brief and whole-limb occlusions, and out-of-distribution limits on pathologies absent from training.
\end{enumerate}

\section{Related Work}

Deterministic monocular biomechanical methods either regress skeletal coordinates directly or optimize a differentiable body model over time. Feedforward regressors such as HSMR \parencite{xia_reconstructing_2025}, BioPose \parencite{koleini_biopose_2025}, and B3D \parencite{cotton_biomechanical_2026} predict joint angles from video, whereas trajectory optimizers including the Portable Biomechanics Laboratory (PBL) \parencite{peiffer_portable_2025}, CAMBR \parencite{cotton_cambr_2026}, and OpenCap Monocular \parencite{gilon_opencap_2026} fit a skeleton to 2D keypoints across time. However, attaching a biomechanical model to a strong regressor did not lower joint-angle error across thirteen monocular pipelines \parencite{li_impact_2025}, and 2D reprojection residuals correlate weakly with true 3D joint error \parencite{peiffer_portable_2025}, leaving clinicians with no warning when self-occlusion or depth ambiguity degrades an estimate. Existing probabilistic pose estimators only partially close this gap: single-frame normalizing flows omit temporal continuity \parencite{kolotouros_probabilistic_2021}, while conditional multi-hypothesis lifters \parencite{shan_diffusion-based_2023, wang_fmpose3d_2026} are evaluated by minimum-of-$K$ error, which rewards miscalibrated distributions \parencite{pierzchlewicz_optimizing_2023}.

Learned generative priors instead steer unconditional motion samples toward observations at test time \parencite{chung2023diffusion, pierzchlewicz_energy_2023}. Building on energy-guided diffusion (EGG) \parencite{pierzchlewicz_energy_2023}, Platypose \parencite{pierzchlewicz_platypose_2024} guided a 3D keypoint diffusion model with a 2D reprojection energy for calibrated monocular pose lifting, while RoHM \parencite{zhang_rohm_2024} conditioned motion diffusion under occlusion and ProjFlow \parencite{watanabe_projflow_2026} projected flow-matching motion samples onto linear spatial constraints. We build on conditional flow matching \parencite{lipman_flow_2023} with a sequence transformer architecture \parencite{peebles_scalable_2023}. With Gaussian noise and a linear interpolant, flow matching is diffusion with a linear noise schedule \parencite{lipman_flow_2023}; we differ in pairing noise and data by minibatch optimal transport \parencite{tong_improving_2024} and in sampling the learned ordinary differential equation (ODE) deterministically. Guiding a deterministic sampler requires care because intermediate states are never re-noised; we combine four recent mechanisms, detailed in the Methods: FlowDPS endpoint recombination \parencite{kim_flowdps_2025}, differentiation through the flow network \parencite{feng_guidance_2025}, guidance scaled to the prior velocity norm \parencite{song_flow_2025}, and Gauss--Newton manifold projection \parencite{utkarsh_physics-constrained_2025}.

\section{Methods}

\begin{figure}[!htbp]
\centering
\includegraphics[width=1\linewidth]{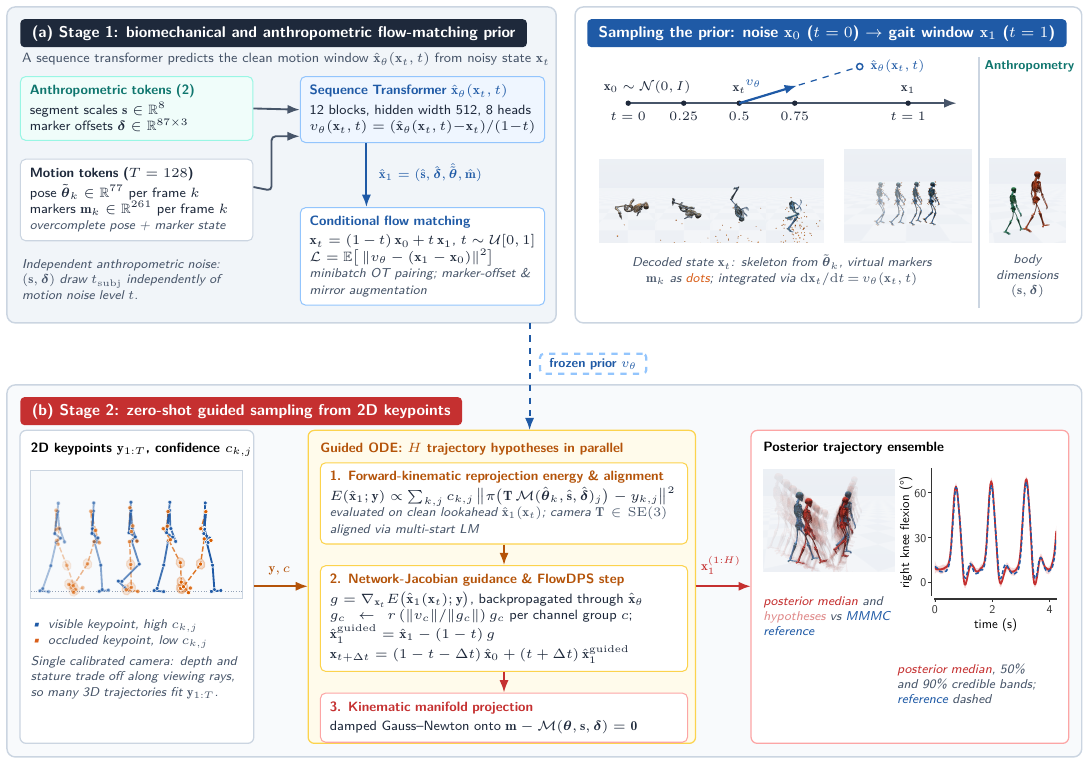}
\caption[]{Overview of the Biomechanical Flow Model. (a) Two anthropometric prefix tokens (segment scales $\mathbf{s}$ and marker offsets $\boldsymbol{\delta}$) and $T$ canonical pose-and-marker tokens are modeled by a sequence transformer trained via conditional flow matching with independent anthropometric noise levels. (b) At test time, an ensemble of trajectory hypotheses is steered toward monocular 2D keypoints via forward-kinematic reprojection gradients backpropagated through the network, FlowDPS endpoint recombination, camera alignment, and kinematic manifold projection.}
\label{fig-overview}
\end{figure}

\subsection{Musculoskeletal model}

We adopt the differentiable 40-DOF MuJoCo \parencite{todorov_mujoco_2012} musculoskeletal model from our multi-camera reconstruction pipeline \parencite{cotton_differentiable_2025, al-hafez_locomujoco_2023}. It comprises 21 rigid segments, a 6-DOF pelvis root, 34 hinge joints, and 87 virtual MoVi anatomical markers \parencite{ghorbani_movi_2021, sarandi_learning_2023}, mapping generalized coordinates and participant anthropometry to 3D marker positions via forward kinematics:
\begin{equation}
\mathbf{m} = \mathcal{M}(\boldsymbol{\theta}, \mathbf{s}, \boldsymbol{\delta}),
\end{equation}
where $\boldsymbol{\theta} \in \mathbb{R}^{41}$ contains root translation, unit-quaternion orientation, and 34 joint angles, $\mathbf{s} \in \mathbb{R}^{8}$ contains segment-scale offsets governing limb lengths and overall stature, and $\boldsymbol{\delta} \in \mathbb{R}^{87 \times 3}$ contains per-marker anatomical offsets. Implemented in MuJoCo-MJX, $\mathcal{M}$ propagates analytical gradients with respect to $\boldsymbol{\theta}$, $\mathbf{s}$, and $\boldsymbol{\delta}$. Throughout, $k \in \{1, \ldots, T\}$ indexes video frames ($T = 128$ at 30 Hz, 4.3 s) and $t \in [0, 1]$ indexes continuous flow time from Gaussian noise ($t = 0$) to clean motion ($t = 1$).

\subsection{Canonical overcomplete state and architecture}

An $\mathrm{SE}(2)$ ground-plane transform canonicalizes each motion window by placing the first-frame pelvis horizontally at the origin with zero heading. Each window forms $T + 2$ tokens of width 338 (Figure~\ref{fig-overview}a): two anthropometric prefix tokens ($\mathbf{s}, \boldsymbol{\delta}$) followed by $T$ motion tokens concatenating a 77D continuous pose vector $\tilde{\boldsymbol{\theta}}_k$ (root translation, 6D root rotation \parencite{zhou_continuity_2018}, and $(\sin\theta, \cos\theta)$ hinge pairs) with 261 virtual 3D marker coordinates $\mathbf{m}_k$:
\begin{equation}
\mathbf{x} = \bigl[\,\mathbf{s},\,\boldsymbol{\delta},\,(\tilde{\boldsymbol{\theta}}_1, \mathbf{m}_1),\,\ldots,\,(\tilde{\boldsymbol{\theta}}_T, \mathbf{m}_T)\,\bigr].
\end{equation}
The markers make the state overcomplete, exposing metric trajectories and pose--marker covariance to the flow loss without calling MuJoCo during training.

A 12-block sequence transformer in JAX/Equinox (hidden width 512, 8 heads) processes the $T + 2$ tokens using token-type linear projections, rotary position embeddings \parencite{su_roformer_2022} on motion tokens, and AdaLN-Zero time conditioning \parencite{peebles_scalable_2023} applied per token, so the two prefix tokens can carry their own flow time (below).

\subsection{Conditional flow matching prior}

Conditional flow matching \parencite{lipman_flow_2023} transports Gaussian noise $\mathbf{x}_0 \sim \mathcal{N}(0, I)$ to clean windows $\mathbf{x}_1$ along the linear interpolant $\mathbf{x}_t = (1 - t)\,\mathbf{x}_0 + t\,\mathbf{x}_1$, $t \sim \mathcal{U}[0, 1]$, regressing the conditional velocity $\mathbf{x}_1 - \mathbf{x}_0$. Minibatch optimal transport pairs noise draws with motion windows via exact linear assignment to shorten transport paths \parencite{tong_improving_2024}. The network predicts the clean window $\hat{\mathbf{x}}_\theta(\mathbf{x}_t, t) \approx \mathbf{x}_1$, which avoids a high-variance target near $t \to 1$ and suits our redundant marker state \parencite{li_back_2025, wang_flowhmr_2026}, and defines $v_\theta(\mathbf{x}_t, t) = (\hat{\mathbf{x}}_\theta - \mathbf{x}_t)/(1 - t)$. Since $v_\theta - (\mathbf{x}_1 - \mathbf{x}_0) = (\hat{\mathbf{x}}_\theta - \mathbf{x}_1)/(1 - t)$, the velocity loss is an $\mathbf{x}_1$ loss weighted by $(1 - t)^{ -2}$, which we cap at $\tau^{ -2}$ ($\tau = 0.05$):
\begin{equation}
\mathcal{L} = \mathbb{E}_{t,\,\mathbf{x}_0,\,\mathbf{x}_1}\Bigl[\min\bigl((1 - t)^{-2}, \tau^{-2}\bigr)\,\|\hat{\mathbf{x}}_\theta(\mathbf{x}_t, t) - \mathbf{x}_1\|^2\Bigr].
\end{equation}

\subsection{Independent anthropometric noise schedule}

If all tokens share one noise level $t$, the many markers over many frames that encode the same segment lengths carry more information about stature than the equally noisy anthropometric tokens, so a standard model reads body size from the markers and ignores $(\mathbf{s}, \boldsymbol{\delta})$. To route stature and segment lengths through $(\mathbf{s}, \boldsymbol{\delta})$ and forward kinematics, we modify training in two ways. First, as in Diffusion Forcing \parencite{chen_diffusion_2024}, each window draws an independent anthropometric time $t_{\mathrm{subj}}$: 1 (clean) with probability 0.3, 0 (pure noise) with probability 0.1, and otherwise $\mathcal{U}[0, 1]$. Motion tokens are independently set to pure noise ($t = 0$) with probability 0.05. The prefix tokens use the same interpolant and $\mathbf{x}_1$ loss at $t_{\mathrm{subj}}$ (standardized, with base noise matched to the training mean and SD and loss weight $\boldsymbol{\sigma}^{ -2}$), and optimal transport pairs noise with the motion tokens only. When motion tokens are pure noise and anthropometry is clean, only $(\mathbf{s}, \boldsymbol{\delta})$ can predict height and limb lengths. At test time we either infer body size from video ($t_{\mathrm{subj}} = t$, one slice of the training draws) or lock $(\mathbf{s}, \boldsymbol{\delta})$ to a known calibration ($t_{\mathrm{subj}} = 1$).

Second, as data augmentation during training, each window replaces its marker offsets $\boldsymbol{\delta}$ with those of a randomly drawn training trial (probability 0.5) and mirrors trajectories sagittally (probability 0.5), recomputing consistent 3D markers via $\mathcal{M}(\boldsymbol{\theta}, \mathbf{s}, \boldsymbol{\delta})$. We train a single model on 9,832 synchronized multi-camera markerless motion capture (MMMC) trials (54 h) from 565 rehabilitation-hospital participants \parencite{cotton_differentiable_2025} under institutional review board approval (200,000 AdamW steps, batch size 128, peak learning rate 10\textsuperscript{-4} with cosine decay) and evaluate it across participant-disjoint splits comprising a matched training-partition subset (``train'', 20 walking and 20 TUG trials from 23 participants withheld from weight updates to match precomputed PBL samples and select sampler settings), a held-out test set (40 walking and 40 TUG trials from 50 unseen participants), and 21 frontal pediatric gait-laboratory videos against optical motion capture (including 7 children with cerebral palsy crouch gait).

\subsection{Guided sampling from monocular keypoints}\label{guided-sampling}

Given undistorted 2D keypoints $y_{k,j}$ and detector confidences $c_{k,j}$ from a single calibrated camera \parencite{cotton_posepipe_2022, sarandi_learning_2023, peiffer_portable_2025}, we steer unconditional trajectories from the prior ODE $\mathrm{d}\mathbf{x}_t/\mathrm{d}t = v_\theta(\mathbf{x}_t, t)$ toward the posterior $p(\mathbf{x}_1 \mid y) \propto p(\mathbf{x}_1)\,\exp( -E(\mathbf{x}_1; y))$, with the reprojection energy $E$ below, by adding a guidance term at test time without retraining (Figure~\ref{fig-overview}b, with $H = 32$ hypotheses integrated in parallel over $N = 200$ steps, the first $N_w = 20$ of them unguided). Because observations depend on clean motion rather than the noisy state $\mathbf{x}_t$, at each flow time $t$ we estimate the clean trajectory in one network pass as the conditional mean $\mathbb{E}[\mathbf{x}_1 \mid \mathbf{x}_t]$, which the $\mathbf{x}_1$ parameterization returns directly:
\begin{equation}
\hat{\mathbf{x}}_1(\mathbf{x}_t) = \mathbf{x}_t + (1 - t)\,v_\theta(\mathbf{x}_t, t) = \hat{\mathbf{x}}_\theta(\mathbf{x}_t, t).
\end{equation}

\textbf{Forward-kinematic reprojection energy.} Decoding $(\hat{\boldsymbol{\theta}}_k, \hat{\mathbf{s}}, \hat{\boldsymbol{\delta}})$ from $\hat{\mathbf{x}}_1(\mathbf{x}_t)$, we score it against the keypoints by forward-kinematic reprojection:
\begin{equation}
E(\hat{\mathbf{x}}_1; y) \propto \sum_{k=1}^{T}\sum_{j=1}^{N_m} c_{k,j}\,\bigl\|\pi\bigl(\mathbf{T}\,\mathcal{M}(\hat{\boldsymbol{\theta}}_k, \hat{\mathbf{s}}, \hat{\boldsymbol{\delta}})_j\bigr) - y_{k,j}\bigr\|^2,
\end{equation}
where $\pi$ is calibrated perspective projection across the $N_m$ anatomical markers and $\mathbf{T} \in \mathrm{SE}(3)$ maps the origin-anchored trajectory into the camera frame. Reprojecting through $\mathcal{M}$ makes the energy directly sensitive to joint angles, segment lengths, and stature. Setting $c_{k,j} = 0$ drops occluded keypoints from the energy.

\textbf{Backpropagating through the network and FlowDPS integration.} Updating $\hat{\mathbf{x}}_1$ along $-\nabla_{\hat{\mathbf{x}}_1} E$ would ignore the correlations the prior has learned, changing each frame only through its own keypoints and occluded limbs not at all. Instead, following Flow Posterior Sampling (FlowDPS) \parencite{kim_flowdps_2025} and covariance-preconditioned guidance \parencite{feng_guidance_2025}, we differentiate $E(\hat{\mathbf{x}}_1(\mathbf{x}_t); y)$ \textit{through} the flow transformer with respect to the noisy state $\mathbf{x}_t$, obtaining the guidance term $g_t(\mathbf{x}_t) = \nabla_{\mathbf{x}_t} E(\hat{\mathbf{x}}_1(\mathbf{x}_t); y)$. Approximating the likelihood of the noisy state by that of its denoised estimate, $p(y \mid \mathbf{x}_t) \approx \exp( -E(\hat{\mathbf{x}}_1(\mathbf{x}_t); y))$ \parencite{chung2023diffusion}, makes $-g_t$ an estimate of the likelihood score $\nabla_{\mathbf{x}_t} \log p(y \mid \mathbf{x}_t)$, which, for independent Gaussian noise, adds to the prior velocity with weight $(1 - t)/t$. Because $\hat{\mathbf{x}}_1$ estimates $\mathbb{E}[\mathbf{x}_1 \mid \mathbf{x}_t]$, the second-order Tweedie identity gives $\partial\hat{\mathbf{x}}_1 / \partial\mathbf{x}_t = t\,(1 - t)^{ -2}\,\mathrm{Cov}[\mathbf{x}_1 \mid \mathbf{x}_t]$ \parencite{efron_tweedies_2011, meng_estimating_2021} (exact for independent Gaussian noise, approximate under the minibatch coupling and for the trained network), so backpropagating through the network preconditions pixel residuals with the prior's cross-joint and temporal covariance and visible keypoints update correlated joints, occluded limbs, and body scales coherently. Within each channel group (scales, offsets, root, joint angles, markers), $g_t$ is scaled to a fraction $r = 0.03$ of the prior velocity norm \parencite{song_flow_2025}, and for scales and offsets further capped at a norm of 0.01. Because this discards the gradient magnitude, the ensemble approximates rather than samples $p(\mathbf{x}_1 \mid y)$; we call its quantiles credible intervals and test their calibration empirically below. Because the camera constrains the clean motion estimate, FlowDPS forms the guided endpoint $\hat{\mathbf{x}}_1^{\mathrm{guided}} = \hat{\mathbf{x}}_1 - (1 - t)\,g_t(\mathbf{x}_t)$, a preconditioned gradient step, and recombines it with the retained noise estimate $\hat{\mathbf{x}}_0 = \mathbf{x}_t - t\,v_\theta(\mathbf{x}_t, t)$ on the linear interpolant:
\begin{equation}
\mathbf{x}_{t+\Delta t} = (1 - t - \Delta t)\,\hat{\mathbf{x}}_0 + (t + \Delta t)\,\hat{\mathbf{x}}_1^{\mathrm{guided}}.
\end{equation}
This equals an Euler step of the prior ODE plus a guidance term $-(t + \Delta t)(1 - t)\,g_t(\mathbf{x}_t)$ whose weight the interpolation schedule sets, not $\Delta t$, so the total guidance grows with $N$.

\textbf{Camera extrinsic alignment.} Because trajectories are anchored at the canonical origin while the camera pose is unknown, each hypothesis solves one stationary rigid transform $\mathbf{T} \in \mathrm{SE}(3)$ at every guided step by refining, via Levenberg--Marquardt, the two lowest-cost of three closed-form initializations (perspective DLT, weak-perspective affine, and apparent-stature depth with calibrated pitch/roll) together with the previous step's solution, detached during backpropagation so gradients move the trajectory rather than the camera.

\textbf{Kinematic manifold consistency.} As in PCFM \parencite{utkarsh_physics-constrained_2025}, every $P = 60$ steps damped Gauss--Newton projects the guided clean estimate $\hat{\mathbf{x}}_1^{\mathrm{guided}}$, not $\mathbf{x}_t$, onto marker--skeleton consistency $h(\mathbf{x}) = \mathbf{m} - \mathcal{M}(\boldsymbol{\theta}, \mathbf{s}, \boldsymbol{\delta}) = 0$ with anthropometric tokens fixed before the recombination re-noises it, and once more at $t = 1$. Each in-loop projection is followed by three steps of PCFM's relaxed constraint correction.

\subsection{Evaluation}\label{evaluation}

We evaluate posterior-median joint-angle mean absolute error (MAE, \ensuremath{^\circ}) over a single consolidated set of 26 reliable skeletal degrees of freedom (bilateral 3-DOF hip, knee, ankle, subtalar, 3-DOF shoulder, and elbow; 3-DOF lumbar and neck), excluding the toe, forearm, and wrist DOFs that shod 2D keypoints do not constrain. Global spatial accuracy is quantified via Procrustes-aligned marker error (PA-MPJPE), pelvis translation RMSE, and horizontal displacement ratio. Following \textcite{cotton_biomechanical_2025} and \textcite{donahue_calibrated_2026}, we test whether the predicted uncertainty scales with the actual error via the Expected Calibration Error (ECE) \parencite{pierzchlewicz_optimizing_2023}, the mean absolute deviation between empirical probability-integral-transform quantiles and a uniform distribution, alongside central-interval coverage from the 32 hypotheses (128 hypotheses raise it by a finite-ensemble 0.02--0.03 at 4.2$\times$ the time). Paired comparisons use participant-cluster bootstrap 95\% confidence intervals. For occlusion we rerun the test set with zero confidence on every keypoint for a centered 1 s window, or on the right leg throughout, and score hidden and visible joints separately.

\section{Results}

All evaluations score our single trained checkpoint against the synchronized multi-camera markerless reference (MMMC) \parencite{cotton_differentiable_2025}.

\subsection{Comparison with deterministic trajectory optimization, cohort accuracy, and calibration}

\begin{figure}[!htbp]
\centering
\includegraphics[width=1\linewidth]{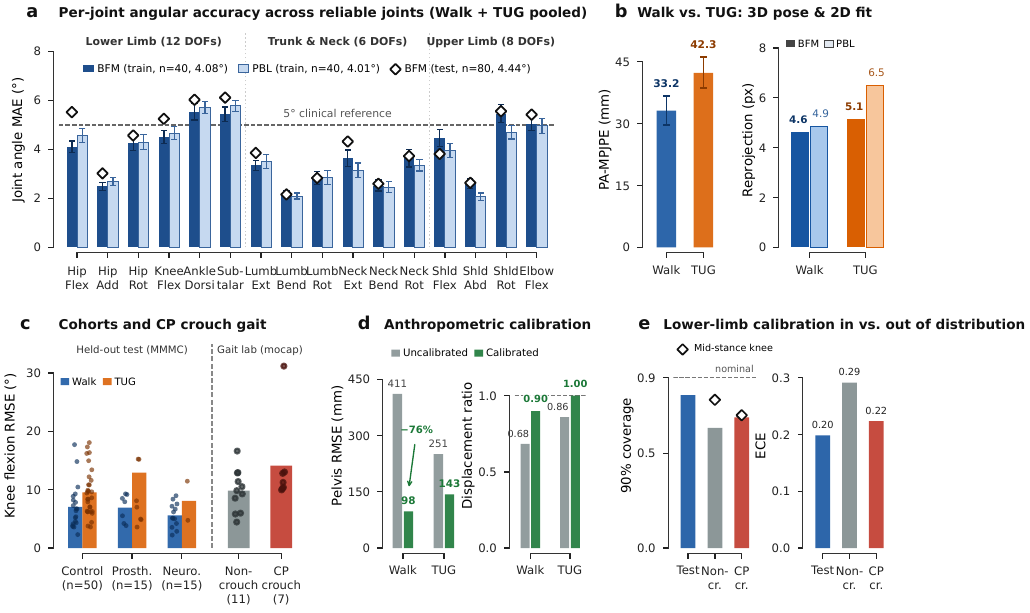}
\caption[]{Cohort accuracy and head-to-head baseline comparison against the multi-camera markerless reference. (a) Per-joint angular MAE across the reliable skeletal joint groups (left and right collapsed into 16 anatomical groups, pooling walking and TUG), comparing BFM and monocular trajectory optimization (PBL \parencite{peiffer_portable_2025}) on the matched train subset ($n = 40$) alongside the held-out test set ($n = 80$). (b) Walking versus TUG 3D pose (PA-MPJPE, test) and 2D reprojection fit (matched train). (c) Knee flexion RMSE on one axis for the held-out clinical cohorts (multi-camera markerless reference) and for gait-laboratory walking without crouch and with out-of-distribution pediatric cerebral palsy crouch gait (optical motion-capture reference). (d) Uncalibrated versus anthropometrically calibrated 3D pelvis RMSE and displacement ratio (test medians). (e) Lower-limb calibration in and out of distribution: empirical 90\% coverage (bars, with diamonds for the mid-stance knee) and ECE on the held-out walking test set and the two gait-laboratory groups.}
\label{fig-accuracy-calibration}
\end{figure}

We first compared BFM head-to-head against the Portable Biomechanics Laboratory (PBL) \parencite{peiffer_portable_2025}, a state-of-the-art deterministic biomechanical trajectory optimizer fitting the identical skeleton to the same monocular 2D keypoints across the matched train subset, on which we also selected the sampler hyperparameters (Figure~\ref{fig-accuracy-calibration}a, Table~\ref{tbl-test-accuracy}). Across the reliable lower-limb, axial, and upper-limb joint groups, BFM matches PBL in joint-angle accuracy across walking and TUG while yielding tighter 2D reprojection fits (Figure~\ref{fig-accuracy-calibration}a,b) and lower hip flexion error during TUG chair transitions (Table~\ref{tbl-test-accuracy}).

\begin{table}
\centering
\caption[]{Held-out test accuracy, uncertainty calibration, and head-to-head comparison against PBL \parencite{peiffer_portable_2025} on the matched train subset against the MMMC reference \parencite{cotton_differentiable_2025} (paired BFM - PBL [95\% CI]) over the reliable skeletal joints.}
\label{tbl-test-accuracy}
{\tiny\setlength{\tabcolsep}{3pt}\resizebox{\ifdim\width>\linewidth\linewidth\else\width\fi}{!}{%
\begin{tabular}{@{}lllll@{}}
\toprule
Metric & Walk test ($n = 40$) & TUG test ($n = 40$) & Walk train: BFM vs PBL ($n = 20$) & TUG train: BFM vs PBL ($n = 20$) \\
\midrule
Joint MAE, mean / median (\ensuremath{^\circ}) & 3.91 / 3.59 & 4.97 / 4.89 & 3.82 vs 3.61 (+0.21 [-0.00, +0.45]) & \textbf{4.34} vs 4.41 (\textbf{-0.07 [-0.31, +0.26]}) \\
Hip / knee / ankle flexion MAE (\ensuremath{^\circ}) & 4.54 / 4.50 / 5.96 & 6.50 / 6.00 / 6.09 & 3.89 / 4.56 / 5.67 vs 3.87 / 4.69 / 5.81 & \textbf{4.32} / \textbf{4.48} / \textbf{5.41} vs 5.25 / 4.64 / 5.61 \\
PA-MPJPE (mm) / Reprojection (px) & 33.2 / 4.0 & 42.3 / 5.3 & 43.8 / 4.6 vs -- / 4.9 & 40.5 / 5.1 vs -- / 6.5 \\
Expected Calibration Error (ECE) & \textbf{0.183} & \textbf{0.137} & 0.154 vs -- & 0.112 vs -- \\
Empirical 50\% / 90\% coverage & \textbf{0.40 / 0.81} & \textbf{0.43 / 0.81} & 0.42 / 0.83 vs -- & 0.44 / 0.81 vs -- \\
\bottomrule
\end{tabular}
}}
\end{table}

\begin{figure}[!htbp]
\centering
\includegraphics[width=1\linewidth]{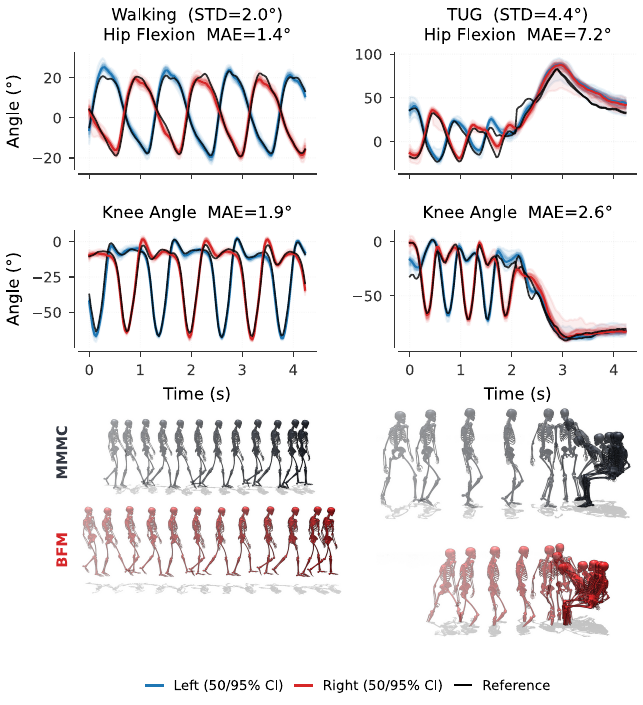}
\caption[]{Lowest-spread held-out test reconstructions for walking (left) and TUG (right). Top two rows: bilateral hip flexion and knee flexion (left blue, right red) showing posterior median (solid), 50\%, 68\%, and 95\% credible bands (shaded), and MMMC reference (black). Bottom row: strobe view every 0.4 s of the MMMC reference (charcoal) and the coordinate-wise posterior median (red), aligned on the ground plane and viewed from the same camera, later frames more opaque.}
\label{fig-confident-reconstruction}
\end{figure}

Across the held-out test set of unseen participants, BFM maintains low joint-angle and 3D positional error on both walking and TUG (Figure~\ref{fig-accuracy-calibration}a,b, Table~\ref{tbl-test-accuracy}). Reconstruction quality generalizes across unseen able-bodied controls, lower-limb prosthesis users, and neurologic gait-clinic participants (Figure~\ref{fig-accuracy-calibration}c), tracking reference joint waveforms and whole-body motion smoothly (Figure~\ref{fig-confident-reconstruction}). Without any post-hoc rescaling the credible intervals are close to calibrated, if slightly narrow: the central 90\% interval covers about four in five reference angles on both tasks (Table~\ref{tbl-test-accuracy}).

\subsection{Resilience to temporal blackouts and whole-limb occlusion}

\begin{figure}[!htbp]
\centering
\includegraphics[width=1\linewidth]{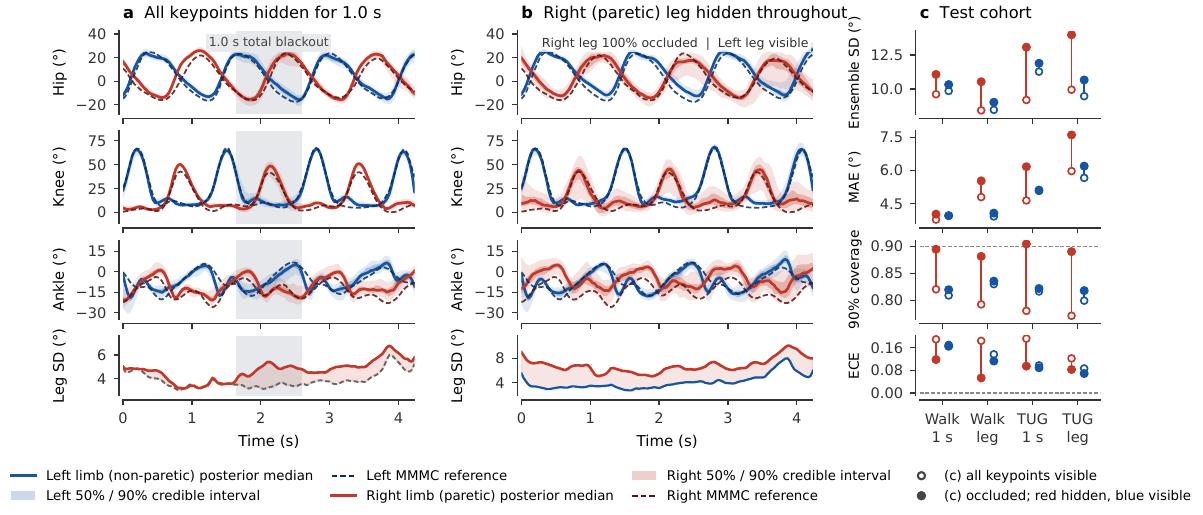}
\caption[]{Guided sampling with hidden keypoints set to zero confidence. (a, b) A held-out hemiparetic walking trial, showing the posterior median, shaded 50\%/90\% credible bands, and dashed MMMC reference for the left (blue) and paretic right (red) limbs. (a) All keypoints hidden for 1.0 s (gray), with the bottom row showing per-frame SD over both legs with (red) and without (gray dashed) the blackout; the excess left after the blackout lies mostly in the ankles and hips of both legs. (b) The right leg hidden throughout, with the bottom row showing per-frame SD of the hidden (red) and visible (blue) leg. (c) All 80 test trials: ensemble SD, MAE, 90\% coverage and ECE on hidden (red) and visible (blue) joints with all keypoints visible (open) and with keypoints hidden (filled), for the 1 s blackout and the hidden right leg on walking and TUG.}
\label{fig2-uncertainty}
\end{figure}

With hidden keypoints at zero confidence, on a held-out hemiparetic walking trial the prior bridges a one-second blackout of every keypoint while keeping the asymmetric flexion and stride phase (Figure~\ref{fig2-uncertainty}a), and reconstructs a hidden paretic leg in antiphase with the visible leg, with wider bands on the hidden side (Figure~\ref{fig2-uncertainty}b). The key question is whether the predicted uncertainty still scales with the error \parencite{donahue_calibrated_2026} where the video carries no information. Across all test trials (Figure~\ref{fig2-uncertainty}c), hiding keypoints raised the spread of the hidden joints proportionally more than their error, so their 90\% intervals moved from covering about four in five reference angles to nearly nine in ten and the ECE fell, while the visible joints were nearly unchanged. Where keypoints are missing, the posterior widens rather than guessing confidently.

\subsection{Anthropometric calibration and monocular scale--depth bias}

Because stature and camera distance trade off along viewing rays, uncalibrated monocular tracking shows a mild scale--depth compression bias: inferring anthropometry from video alone favors a slightly smaller skeleton closer to the camera, compressing horizontal displacement ratios (Figure~\ref{fig-accuracy-calibration}d) while preserving internal joint angles. Supplying individual anthropometric calibration alongside camera intrinsics resolves this ambiguity (Figure~\ref{fig-accuracy-calibration}d), removing three quarters of walking 3D pelvis translation error and restoring displacement ratios near unity. Constraining overall stature from a simple standing height measurement represents a natural clinical extension.

\subsection{Out-of-distribution pediatric crouch gait}

Frontal walking videos from a pediatric gait laboratory, scored against optical motion capture, test behavior beyond the training support. Children with cerebral palsy and crouch gait, a pathology absent from training, are not reconstructed well: the prior pulls the flexed mid-stance knee toward its training cohort's more extended knee (Figure~\ref{fig-accuracy-calibration}c). Calibration also degrades in this laboratory for crouch and non-crouch participants alike, because error roughly doubles while the spread grows only modestly (Figure~\ref{fig-accuracy-calibration}e). Part of this likely reflects the change to a marker-based reference. The spread therefore neither singles out the crouch trials nor covers their knee bias.

\subsection{Guided sampler ablations}

The ablations (Table~\ref{tbl-ablations}) show which sampler mechanisms matter. Backpropagating the reprojection energy through the flow network is indispensable: differentiating only with respect to $\hat{\mathbf{x}}_1$, with the same $(1 - t)$ guidance taper, loses the prior's cross-joint covariance and fails to track the keypoints. Total guidance matters next: an Euler step, which applies 1/36 of the FlowDPS total, loses 4.7\ensuremath{^\circ}, but equals the full sampler once $r = 1.1$ matches that total, so FlowDPS helps through guidance strength, not noise recombination. Guiding the anthropometric tokens is modestly better than leaving them unguided. Projection does not change accuracy or calibration, but it keeps the generated markers on the skeleton.

\begin{table}
\centering
\caption[]{Guided-sampler ablations (matched train subset, walking and TUG pooled, reliable joints). $\Delta$ MAE: paired difference from the full sampler [95\% CI, participant clusters]. Reprojection: mean pixel error of the forward-kinematic markers. KD: mean marker-to-skeleton discrepancy (mm). Each row changes one setting; $r = 0.03$ unless stated.}
\label{tbl-ablations}
{\tiny\setlength{\tabcolsep}{3pt}\resizebox{\ifdim\width>\linewidth\linewidth\else\width\fi}{!}{%
\begin{tabular}{@{}llllll@{}}
\toprule
Sampler configuration & Joint MAE (\ensuremath{^\circ}) & $\Delta$ MAE [95\% CI] (\ensuremath{^\circ}) & ECE & Reprojection (px) & KD (mm) \\
\midrule
\textbf{Full (forward-kinematic guidance, projection every 60 steps)} & \textbf{4.08} & --- & 0.133 & 4.9 & \textbf{0.0} \\
Gradient w.r.t. $\hat{\mathbf{x}}_1$ (no network Jacobian) & 14.89 & +10.81 [+9.56, +12.07] & 0.241 & 69.2 & 0.0 \\
Euler step, guidance weighted by $\Delta t$ (summed weight 0.90 vs 32.85) & 8.78 & +4.69 [+3.84, +5.68] & 0.088 & 90.8 & 0.0 \\
Euler step with $r = 1.1$ (summed weight matched) & 4.14 & +0.06 [-0.02, +0.13] & 0.131 & 6.0 & 0.0 \\
Fewer integration steps ($N = 20$, $N_w = 2$, $P = 6$; less total guidance) & 7.43 & +3.35 [+2.61, +4.24] & 0.098 & 32.1 & 0.0 \\
No guidance on anthropometric tokens & 4.27 & +0.19 [+0.11, +0.29] & 0.105 & 5.6 & 0.0 \\
No projection (periodic or final) & 4.09 & +0.01 [-0.01, +0.03] & 0.129 & 4.3 & 22.9 \\
Guiding the virtual markers directly instead of through forward kinematics & 4.00 & -0.08 [-0.19, -0.00] & 0.201 & 6.8 & 0.0 \\
\bottomrule
\end{tabular}
}}
\end{table}

Although the virtual markers are needed during training, guidance need not steer them: guiding them directly matches forward-kinematic guidance in joint-angle accuracy, if anything slightly better, but fits the keypoints worse and yields a narrower, overconfident ensemble (Table~\ref{tbl-ablations}), so pure forward-kinematic guidance is our default.

\section{Discussion}

We trained a statistical flow-matching prior over anatomical joint-angle trajectories and participant anthropometry and benchmarked it head-to-head against a state-of-the-art deterministic biomechanical trajectory optimizer \parencite{peiffer_portable_2025}. Across unseen clinical assessments, guided flow sampling matched deterministic optimization in joint-angle accuracy while adding calibrated credible intervals \parencite{donahue_calibrated_2026} and resilience to temporal and whole-limb occlusions. Unlike Platypose \parencite{pierzchlewicz_platypose_2024} and flow-matching lifters \parencite{wang_fmpose3d_2026}, which operate on unconstrained 3D keypoint positions, BFM yields physically consistent joint trajectories, optional anthropometric calibration, and marginal variances suitable for downstream physics tracking \parencite{cotton_kintwin_2025}.

Our ablations and training design isolate what made biomechanical flow matching work reliably. First, with clean-window prediction \parencite{li_back_2025} and velocity-normalized FlowDPS updates \parencite{kim_flowdps_2025, song_flow_2025}, we observed smooth joint waveforms without Neural ODE regularization or post-hoc low-pass filtering, though no ablation isolates either component's effect on smoothness. Second, differentiating forward-kinematic reprojection \textit{through} the flow network \parencite{feng_guidance_2025} proved indispensable so 2D residuals update correlated joints, occluded limbs, and body dimensions coherently. Third, whereas 3D virtual markers in the training state were critical for learning metric scale and pose--marker covariance without calling MuJoCo during training, guidance works best through forward kinematics alone, since steering the markers directly kept accuracy but made the ensemble overconfident. Periodic projection \parencite{utkarsh_physics-constrained_2025} only keeps markers and skeleton consistent. Fourth, independent anthropometric noise levels and marker-offset augmentation prevented stature shortcuts from noisy markers, enabling anthropometric conditioning that removes scale--depth compression.

Several limitations point to immediate extensions. First, our alignment assumes calibrated intrinsics and a stationary camera; because $\mathbf{T}$ simply maps the ground frame into the camera frame, per-frame camera poses $\mathbf{T}_k$ from visual-geometry estimators \parencite{shen_gvhmr_2024} should extend BFM directly to moving cameras. Second, reconstruction quality remains bounded by upstream 2D keypoint detections and confidences. Third, our reference is single-center markerless capture, and the unflagged failure on pediatric crouch gait shows that the prior reflects only its training support, motivating multi-site pediatric and spastic cohorts and out-of-distribution detection. Fourth, BFM is trained on 128-frame (4.3 s) windows, so longer recordings are windowed independently; per-frame noise levels conditioned on the previous window, as in diffusion forcing \parencite{chen_diffusion_2024}, would allow rollout of any length. Finally, guidance takes tens of seconds per window; few-step distillation \parencite{wang_fmpose3d_2026} could accelerate it.

\section{Conclusion}

Steering a continuous-time flow-matching prior over anatomical joint-angle trajectories and participant anthropometry enables accurate monocular 3D biomechanical reconstruction with calibrated uncertainty. Across unseen participants, BFM matches state-of-the-art deterministic trajectory optimization in joint-angle accuracy while providing credible intervals that widen where keypoints are hidden, keeping their coverage near nominal, bridging whole-body and unilateral limb occlusions, and resolving monocular scale--depth compression when anthropometry is supplied. Biased, overconfident predictions on out-of-distribution pediatric crouch gait simultaneously show that trustworthy clinical deployment requires pairing calibrated generative inference with pathologically diverse movement priors.

\section*{Acknowledgments}
\small
R.J.C. is supported by the National Institute of Child Health and Human Development and the National Center for Medical Rehabilitation Research (NIH R01HD114776) and the National Institute on Disability, Independent Living, and Rehabilitation Research (NIDILRR grant number 90REGE0030), and the Research Accelerator Program of the Shirley Ryan AbilityLab. F.H.S. is supported by the German Research Foundation (DFG): SFB 1233 (Robust Vision: Inference Principles and Neural Mechanisms, Project-ID 276693517) and by the European Research Council (ERC) under the European Union's Horizon Europe research and innovation program (Grant agreement No. 101171526).\par
\smallskip\noindent AI-assisted coding and research agents, including Claude (Anthropic) and Google Antigravity, were used for code development, running experiments, and assistance with manuscript preparation. The manuscript was extensively reviewed and verified by the authors, who take full responsibility for its content.\par
\normalsize

\clearpage
\section{Supplementary Material}

This supplement expands the Methods with the model and training details, the full guided sampler, what each sampler ablation changes, and how calibration is scored. It reports no results beyond the main paper. Time follows the main paper, with $t = 0$ noise and $t = 1$ data.

\subsection{S1. Model and training}

\textbf{Anthropometry and state.} Segment $b$ is scaled by $1 + \sum_k M_{bk} s_k$ for a fixed binary matrix $M$ that maps the 8 scale offsets to the 21 segments; the overall scale $s_7$ applies to every segment. The 269 active anthropometric entries (8 scales, 261 marker offsets) are zero-padded to the token width of 338, and the padding carries no loss. These entries vary far less across the cohort than the unit-variance motion channels, so under a common noise schedule they would be far noisier relative to their signal. The network therefore z-scores them with training-cohort statistics on input and maps its predictions back on output. It weights their loss by $1/\sigma_j^2$ and draws their base noise from a Gaussian matched to the cohort mean and variance.

\textbf{Network.} The 12 transformer blocks (JAX/Equinox \parencite{kidger_equinox_2021}) have 8 heads, width 512 and 2048-wide GELU feed-forward layers. The flow time passes through a sinusoidal encoding \parencite{vaswani_attention_2017} and a two-layer MLP, and modulates every block and the final layer norm through AdaLN-Zero \parencite{peebles_scalable_2023}. Every token receives its own time, which lets the subject tokens run on a separate clock. The rotary position tables of the motion tokens \parencite{su_roformer_2022} start at the standard values but are trained, so they end as a learned per-position modulation rather than an exact rotation. The two subject tokens receive learned type embeddings instead.

\textbf{Clean-sample prediction.} The network predicts the clean window, and the velocity is $(\hat{\mathbf{x}}_\theta - \mathbf{x}_t)/(1 - t)$. The velocity loss weight $\min(1, (1 - t)/\tau)^2$ with $\tau = 0.05$ is equivalent to weighting the clean-sample error by $1 / \max(1 - t, \tau)^2$, which caps the weight where the velocity target becomes singular.

\textbf{The subject clock.} Under one shared noise level, the 128 frames of noisy markers reveal stature before the equally noisy subject tokens do, and the network learns to ignore those tokens. We conditioned 128 unconditional samples on the clean subject tokens of 128 real held-out windows and correlated the overall scale with the standing pelvis height of each sample.

\begin{itemize}
\item Under a shared clock the correlation was 0.05. Z-scoring the tokens left it at 0.04, and also matching their base noise raised it only to 0.29.
\item With the separate subject clock it rose to 0.66, against 0.72 in the real windows. The sample heights correlated 0.74 with the same participants' real heights, with a bias of +3 mm.
\end{itemize}

During training the subject time is 1 (clean) with probability 0.3, 0 with probability 0.1, and otherwise uniform; the motion time is independently set to 0 with probability 0.05. At test time, inferring anthropometry uses one shared time for all tokens, and known anthropometry is held at $t = 1$.

\textbf{Training.} Each 128-frame training window first takes the marker offsets of a randomly drawn training trial (probability 0.5). It is then mirrored through the sagittal plane (probability 0.5), which swaps the left and right joint and scale channels and negates the coronal and transverse spine and neck angles. The markers are recomputed through $\mathcal{M}$, so every token satisfies $\mathbf{m}_k = \mathcal{M}(\boldsymbol{\theta}_k, \mathbf{s}, \boldsymbol{\delta})$ exactly. Within each minibatch an exact linear assignment on the motion tokens pairs noise with data \parencite{tong_improving_2024}. The subject-token noise is never permuted, so it stays independent of the data.

AdamW \parencite{loshchilov_decoupled_2019} ($\beta = (0.9, 0.999)$, $\epsilon = 10^{ -8}$, weight decay 0.01) runs 200,000 steps at batch size 128. The peak learning rate of 10\textsuperscript{-4} follows 1,000 linear warm-up steps and decays on a cosine to zero at step 199,000. Gradients are clipped at norm 1, and there is no dropout or auxiliary regularizer. The loader replays a fixed stream of 20,000 augmented batches, so training makes ten passes over the same 2.56 M windows with fresh noise, times and pairings. We report the raw weights; the exponential moving average (0.999) differs by at most 0.01\ensuremath{^\circ} in development-set error.

\textbf{Trial selection.} Evaluation trials were drawn in a fixed random order from multi-camera recordings that passed these quality gates:

\begin{itemize}
\item multi-camera reprojection error of at most 10 px;
\item trial-mean keypoint confidence of at least 0.3 (0.5 for test walking);
\item for walking, at least 5 m of net pelvis displacement;
\item a 128-frame window whose camera had a mean keypoint confidence of at least 0.3, and, for walking, a pelvis speed of at least 0.5 m/s at every frame.
\end{itemize}

Windows and cameras were fixed before any monocular reconstruction. Keypoints were detected on person boxes from the laboratory's multi-camera tracker; a deployed system would need a single-camera tracker.

\subsection{S2. Guided sampler}

Each window runs 32 hypotheses, in batches of 16, for $N = 200$ steps, the first $N_w = 20$ unguided; Algorithm 1 gives the full loop.

\textbf{Camera alignment.} Each hypothesis solves one stationary $\mathbf{T} \in \mathrm{SE}(3)$ at every guided step. Because a single local solve can settle in a wrong yaw basin, three closed-form starts are formed:

\begin{enumerate}
\item a perspective direct linear transform \parencite{hartley_multiple_2004}, projected onto $\mathrm{SO}(3)$;
\item a weak-perspective affine solve, well conditioned early, while the pose is noisy;
\item an apparent-size depth start from the ratio of 3D to 2D marker spread, with the rotation from the calibrated camera tilt.
\end{enumerate}

The two lowest-cost starts and the previous step's solution each receive 15 Levenberg--Marquardt iterations \parencite{levenberg_method_1944, marquardt_algorithm_1963}, and the best is kept. $\mathbf{T}$ is held fixed while the energy is differentiated.

\textbf{Energy and step size.} The energy divides the confidence-weighted squared pixel residual by $s = \max(c_x, c_y, 128)$, with $(c_x, c_y)$ the principal point. Hidden keypoints in the occlusion experiments have zero confidence. Following \textcite{song_flow_2025}, the guidance term is rescaled within each of five channel groups $c$ (body scales, marker offsets, root, joint angles, markers), with norms pooled over the frames of the window:
\begin{equation}
g_c^{\mathrm{scaled}} = r\,\frac{\|v_c\|}{\|g_c\|}\,g_c, \qquad r = 0.03 .
\end{equation}
The two subject-token groups are each further capped at a norm of 0.01. Because the rescaling discards the gradient magnitude, $s$ affects only the reported energy. Since $\hat{\mathbf{x}}_1 = \mathbf{x}_t + (1 - t)\,v$, a velocity perturbation moves the clean estimate by $(1 - t)$ times as much. This gives $\hat{\mathbf{x}}_1^{\mathrm{guided}} = \hat{\mathbf{x}}_1 - (1 - t)\,g^{\mathrm{scaled}}$, which anneals the guidance as the integration converges.

\textbf{FlowDPS integration.} The noise estimate $\hat{\mathbf{x}}_0 = \mathbf{x}_t - t\,v$ comes from the unguided velocity and is recombined with the guided clean estimate on the interpolant \parencite{kim_flowdps_2025}. Substituting both estimates,
\begin{equation}
\mathbf{x}_{t+\Delta t} = \mathbf{x}_t + \Delta t\,v - (t + \Delta t)(1 - t)\,g^{\mathrm{scaled}},
\end{equation}
an Euler step of the prior plus a guidance term whose weight does not shrink with $\Delta t$. Summed over the 180 guided steps this weight is 32.85; a plain Euler step, which weights the guidance by $\Delta t$, sums to 0.90. Because $r$ multiplies every step's guidance, an Euler step with $r = 0.03 \times 32.85 / 0.90 \approx 1.1$ applies the same total, though spread evenly over the steps instead of peaking at mid-integration as $(t + \Delta t)(1 - t)$ does.

\textbf{Projection onto the forward-kinematic manifold.} The markers and joint angles must satisfy $h(\mathbf{x}) = \mathbf{m} - \mathcal{M}(\boldsymbol{\theta}, \mathbf{s}, \boldsymbol{\delta}) = 0$. Starting from $\hat{\mathbf{x}}_1^{\mathrm{guided}}$, damped Gauss--Newton updates each frame's 338-D motion token (pose and markers jointly), with the subject tokens held fixed:
\begin{equation}
\hat{\mathbf{x}}_1 \leftarrow \hat{\mathbf{x}}_1 - J^\top\bigl(JJ^\top + \lambda\,\mathrm{diag}(\mathrm{diag}(JJ^\top)) + \epsilon I\bigr)^{-1} h(\hat{\mathbf{x}}_1),
\end{equation}
with $J = \partial h / \partial \hat{\mathbf{x}}_1$, $\lambda = 10^{ -3}$ and $\epsilon = 10^{ -6}$ \parencite{marquardt_algorithm_1963}.

\begin{itemize}
\item \textbf{When it runs:} at steps 0, 60, 120 and 180 (the first in the warm-up) with up to 5 iterations, and after the last step with up to 20. A frame stops on convergence or once an iteration has grown its residual by more than half.
\item \textbf{Relaxed correction:} on projection steps, three gradient steps (step size 10\textsuperscript{-3}) on $\|u - \mathbf{x}_{t+\Delta t}\|^2 + 10\,\|h(\hat{\mathbf{x}}_1(u))\|^2$ also pull the next state toward the manifold \parencite{utkarsh_physics-constrained_2025}.
\item \textbf{Choice of interval:} $P = 60$ was carried over from a 100-step configuration with the same four projections. On the matched train subset, $P = 20$ or 30 changes the 22-joint MAE by at most 0.03\ensuremath{^\circ} and the 90\% coverage by at most 0.003. Projecting at every step raises the MAE by 0.37\ensuremath{^\circ} (walking, 95\% CI 0.10 to 0.69) and 0.36\ensuremath{^\circ} (TUG, 0.14 to 0.62), narrows the spread by 1.1--1.3\ensuremath{^\circ}, and takes 8--10 times as long as $P = 20$.
\end{itemize}

Inference-time projection trades constraint satisfaction against distributional accuracy \parencite{baldan_physics_2026}. Here its role is to keep the markers on the skeleton.

\begin{algorithm}[t]
\caption{Guided flow matching from monocular keypoints (one hypothesis)}
\label{alg:guided-flow}
\small
\begin{algorithmic}[1]
\Require keypoints $\mathbf{y}$, confidences $c$, intrinsics $\mathbf{K}$, energy $E$, constraint $h$, velocity $v_\theta$, $N$, $N_w$, $P$, $r$
\State Draw $\mathbf{x}_0$ ($\mathcal{N}(0, I)$ on motion tokens, cohort-matched Gaussian on subject tokens); $\Delta t \gets 1/N$
\For{$n = 0, \ldots, N{-}1$}
    \State $t \gets n\,\Delta t$; $v \gets v_\theta(\mathbf{x}_t, t)$; $\hat{\mathbf{x}}_1 \gets \mathbf{x}_t + (1{-}t)\,v$; $\hat{\mathbf{x}}_0 \gets \mathbf{x}_t - t\,v$
    \If{$n \geq N_w$}
        \State Solve $\mathbf{T}$ from $\hat{\mathbf{x}}_1$ and $\mathbf{y}$ (multi-start) and hold it fixed
        \State $g \gets \nabla_{\mathbf{x}_t} E\bigl(\hat{\mathbf{x}}_1(\mathbf{x}_t)\bigr)$ through $v_\theta$; per group $g_c \gets r\,(\|v_c\| / \|g_c\|)\,g_c$, subject groups capped at 0.01
        \State $\hat{\mathbf{x}}_1 \gets \hat{\mathbf{x}}_1 - (1{-}t)\,g$
    \EndIf
    \If{$n \bmod P = 0$}
        \State Project $\hat{\mathbf{x}}_1$ onto $h = 0$ (Gauss--Newton, subject tokens fixed)
    \EndIf
    \State $\mathbf{x}_{t+\Delta t} \gets (1{-}t{-}\Delta t)\,\hat{\mathbf{x}}_0 + (t{+}\Delta t)\,\hat{\mathbf{x}}_1$ \Comment{FlowDPS}
    \If{$n \bmod P = 0$}
        \State Relaxed constraint correction of $\mathbf{x}_{t+\Delta t}$
    \EndIf
\EndFor
\State \textbf{return} $\mathbf{x}_1$ after a final Gauss--Newton projection
\end{algorithmic}
\end{algorithm}
\subsection{S3. What each sampler ablation changes}

Each row of Table~\ref{tbl-ablations} reruns the production sampler on the matched train subset (20 walking and 20 TUG trials) with the same weights, seeds and 32 hypotheses, and changes one setting:

\begin{itemize}
\item \textbf{No network Jacobian.} The energy is differentiated with respect to $\hat{\mathbf{x}}_1$ instead of $\mathbf{x}_t$, keeping the same per-group rescaling and $(1 - t)$ taper. Each frame then moves only through its own keypoints, and unobserved limbs and the subject tokens receive no correlated update.
\item \textbf{Euler step.} The guided clean estimate is turned back into a velocity, $(\hat{\mathbf{x}}_1^{\mathrm{guided}} - \mathbf{x}_t)/(1 - t) = v - g^{\mathrm{scaled}}$, and the state advances by $\Delta t$ times that.

\begin{itemize}
\item The prior step is unchanged, but the guidance weight falls from $(t + \Delta t)(1 - t)$ to $\Delta t$, and the in-loop projection displacement is likewise scaled by $\Delta t/(1 - t)$.
\item At the production $r = 0.03$ the row therefore measures the guidance weight. A second row raises $r$ to 1.1, which matches the FlowDPS total, and so isolates the recombination and the per-step schedule.
\item The same 40 trials were also run with Euler at $r = 0.3$ and $r = 3.0$ (Table~\ref{tbl-supp-euler-r}). Error is lowest at the matched weight and rises on both sides. With too little guidance the ensemble stays wide and its errors are large. With too much, the ensemble is too narrow: its 90\% interval covers the truth only 77\% of the time, and the ECE rises.
\item At the matched weight the Euler sampler is within 0.06\ensuremath{^\circ} and 0.002 ECE of FlowDPS. Its reprojection error is 6.0 px instead of 4.9 px, which is consistent with the evenly spread guidance and the in-loop projection displacement weighted by $\Delta t/(1 -t)$.
\end{itemize}

\item \textbf{Fewer integration steps.} $N = 20$, $N_w = 2$ and $P = 6$, which keeps four in-loop projections. The summed guidance weight falls to 3.71, so this row mixes discretization with total guidance.
\item \textbf{No guidance on anthropometric tokens.} The subject-token cap is set to 0, so the prior alone infers body scale and marker offsets, conditioned through the network on the guided motion.
\item \textbf{No projection.} The in-loop and final projections and the relaxed correction are all off.
\item \textbf{Direct markers.} The energy and the camera alignment use the generated marker channels instead of the forward-kinematic markers ($E = (1 - \alpha)E_{\mathrm{FK}} + \alpha E_{\mathrm{direct}}$ with $\alpha = 1$; the paper uses $\alpha = 0$).
\end{itemize}

\begin{table}
\centering
\caption[]{Euler-step sampler at four guidance scales $r$ against the full FlowDPS sampler, on the same 40 trials as the sampler ablations. The summed weight is in units of the production $r = 0.03$; $r = 1.1$ matches FlowDPS. $\Delta$ MAE: paired difference from the full sampler [95\% CI, participant clusters]. Coverage: 90\% central interval.}
\label{tbl-supp-euler-r}
{\tiny\setlength{\tabcolsep}{3pt}\resizebox{\ifdim\width>\linewidth\linewidth\else\width\fi}{!}{%
\begin{tabular}{@{}llllll@{}}
\toprule
Euler $r$ & 0.03 & 0.3 & 1.1 & 3.0 & FlowDPS, $r = 0.03$ \\
\midrule
Summed weight, in units of $r = 0.03$ & 0.90 & 9.0 & 33 & 90 & 32.85 \\
Joint MAE (\ensuremath{^\circ}) & 8.78 & 5.11 & 4.14 & 4.50 & 4.08 \\
$\Delta$ MAE [95\% CI] (\ensuremath{^\circ}) & +4.69 [+3.84, +5.68] & +1.03 [+0.69, +1.45] & +0.06 [-0.02, +0.13] & +0.42 [+0.21, +0.58] & --- \\
ECE & 0.088 & 0.094 & 0.131 & 0.224 & 0.133 \\
90\% coverage & 0.84 & 0.90 & 0.83 & 0.77 & 0.82 \\
\bottomrule
\end{tabular}
}}
\end{table}

\subsection{S4. Calibration and metrics}

\textbf{Joint sets.} Joint-angle MAE and coverage use the 26 reliable joints. These are, on both sides, hip flexion, adduction and rotation, knee and ankle flexion, subtalar inversion, shoulder flexion, adduction and rotation, and elbow flexion, plus the three lumbar and three neck angles. This excludes in-shoe toe flexion, forearm pronation--supination and wrist flexion and deviation. The ECE uses the 22 clinical joints: hips (2 $\times$ 3), knees, ankles, lumbar spine (3), neck (3), shoulder flexion and adduction (2 $\times$ 2) and elbows.

\textbf{Expected calibration error.} Calibration is scored in joint-angle space \parencite{donahue_calibrated_2026}, adapting the ECE of \textcite{pierzchlewicz_optimizing_2023}. At each frame, every hypothesis and the reference are reduced to their mean absolute distance from the coordinate-wise ensemble median $\mu$ over the 22 joints, $d(\mathbf{x}) = \frac{1}{22}\sum_j |x_j - \mu_j|$. The probability integral transform of the reference is
\begin{equation}
\hat{u} = \frac{1}{H}\sum_{h=1}^{H}\mathbf{1}\bigl[d(\hat{\mathbf{x}}^{(h)}) \le d(\mathbf{x}^{\mathrm{ref}})\bigr],
\end{equation}
which is uniform when the reference is exchangeable with the hypotheses. Pooling the $N$ values over all frames and trials,
\begin{equation}
\mathrm{ECE} = \frac{1}{N}\sum_{i=1}^{N}\bigl|\hat{u}_{(i)} - \tfrac{i - 1}{N}\bigr| .
\end{equation}
An ensemble that is too narrow leaves the reference beyond most hypotheses, which piles $\hat{u}$ up near 1 and raises the ECE. Because each frame is reduced to one distance, errors on different joints can offset each other. Coverage scores each joint separately.

\textbf{Coverage and the finite ensemble.} At each frame and joint, the central $p$ interval runs between the empirical $(1 - p)/2$ and $(1 + p)/2$ quantiles of the $H = 32$ hypotheses, interpolating linearly between order statistics. Coverage is the fraction of reference angles inside it over a trial's frames and joints, averaged over trials.

Even a perfectly calibrated finite ensemble falls short of nominal: a 90\% interval from 32 hypotheses rests on about three hypotheses per tail. For $H$ exchangeable draws, the interpolated interval covers on average $p\,(H - 1)/(H + 1)$ of new values. At 90\% this is 0.85 for $H = 32$ and 0.89 for $H = 128$; at 50\% it is 0.47 and 0.49. The raw coverages of the main paper should therefore be read against 0.85 and 0.47.

Rerunning the 80 test trials with all keypoints and under both occlusion regimes with 128 hypotheses changed three things:

\begin{itemize}
\item the 90\% coverage rose by 0.02--0.03 in every subset, close to this finite-sample prediction;
\item the error changed by -0.19\ensuremath{^\circ} to 0.00\ensuremath{^\circ}, and the pooled ECE by at most 0.03;
\item the cost was time, not memory: on one idle L40S GPU, 70 s instead of 16.5 s per window (4.2$\times$), at 2.7 and 2.8 GiB peak memory.
\end{itemize}

No interval is rescaled anywhere in the paper.

\textbf{Other metrics.} MAE compares the coordinate-wise median of the hypotheses with the reference, wrapping angle differences. Spread is the standard deviation across hypotheses, averaged over frames and joints. PA-MPJPE applies per-frame Procrustes alignment with scale to the 87 forward-kinematic markers of hypothesis $H/2$. Reprojection error is the confidence-weighted mean pixel distance between projected forward-kinematic markers and keypoints, as a per-trial median over hypotheses. The kinematic discrepancy (KD) is the mean $\|\mathbf{m} - \mathcal{M}(\boldsymbol{\theta}, \mathbf{s}, \boldsymbol{\delta})\|$. Pelvis RMSE rigidly aligns each hypothesis's pelvis path without scale. The displacement ratio places each hypothesis in the laboratory frame through its solved camera transform and divides its net pelvis displacement by the reference value. Both pelvis metrics are medians over hypotheses and trials. Paired comparisons bootstrap the per-trial differences over participants (10,000 draws), because trials of one participant are not independent.

{\small
\printbibliography
}

\end{document}